\documentclass[11pt]{article}

\usepackage{acl}

\usepackage{times}
\usepackage{latexsym}
\usepackage[T1]{fontenc}
\usepackage{microtype}
\usepackage{inconsolata}
\usepackage{graphicx}
\usepackage{subcaption}
\usepackage{amsmath}
\usepackage{cleveref}
\usepackage{amssymb}
\usepackage{enumitem}
\usepackage{booktabs}
\usepackage{multirow}
\usepackage{xspace}
\usepackage{todonotes}

\newcommand{\multirowcell}[1]{\begin{tabular}[c]{@{}c@{}}#1\end{tabular}}
\newcommand{\thinkbooster}{\textsc{ThinkBooster}\xspace}
\newcommand{\ecoparagraph}[1]{\noindent\textbf{#1 \;}\ }

\title{ThinkBooster: A Unified Framework for \\ Seamless Test-Time Scaling of LLM Reasoning}
\author{First Author \\ MBZUAI, ETH Zurich, Independent Researcher \\ \texttt{email@mbzuai.ac.ae}}
\author{
\bf Vladislav Smirnov\textsuperscript{1}\enspace
\bf Chieu Nguyen\textsuperscript{1}\enspace
\bf Sergey Senichev\textsuperscript{7}\enspace
\bf Minh Ngoc Ta\textsuperscript{1} \\
\bf Ekaterina Fadeeva\textsuperscript{2}\enspace 
\bf Artem Vazhentsev\textsuperscript{1} \enspace 
\bf Daria Galimzianova\textsuperscript{1}\enspace
\bf Nikolai Rozanov\textsuperscript{1,3} \\
\bf Viktor Mazanov\textsuperscript{6}\enspace 
\bf Jingwei Ni\textsuperscript{2}\enspace 
\bf Tianyi Wu\textsuperscript{4}\enspace 
\bf Igor Kiselev\textsuperscript{5}\enspace 
\bf Mrinmaya Sachan\textsuperscript{2}  \\
\bf Iryna Gurevych\textsuperscript{1} \enspace 
\bf Preslav Nakov\textsuperscript{1}\enspace 
\bf Timothy Baldwin\textsuperscript{1}\enspace 
\bf Artem Shelmanov\textsuperscript{1}
\\
\textsuperscript{1}MBZUAI \;
\textsuperscript{2}ETH Z\"urich \;
\textsuperscript{3}Imperial College London \; 
\textsuperscript{4}NUS \\
\textsuperscript{5}Accenture \;
\textsuperscript{6}Innopolis University \;
\textsuperscript{7}Independent Researcher
\\ \texttt{\{}
\href{mailto:vladislav.smirnov@mbzuai.ac.ae}{\texttt{vladislav.smirnov,}} 
\href{mailto:timothy.baldwin@mbzuai.ac.ae}{\texttt{timothy.baldwin,}} 
\href{mailto:artem.shelmanov@mbzuai.ac.ae}{\texttt{artem.shelmanov}}
\texttt{\}@mbzuai.ac.ae}
}
\begin{document}
\maketitle

% \begin{abstract}
% TODO:
% % TODO: Add abstract summarizing the ThinkBooster framework for seamless test-time compute scaling of reasoning.
% \end{abstract}

\begin{abstract}
Test-time compute (TTC) scaling has emerged as a powerful paradigm for improving large language model (LLM) reasoning by allocating additional compute during inference, e.g., via multi-sample generation and verifier-based reranking. Existing TTC scaling strategies and reasoning scorers remain fragmented, evaluated under inconsistent protocols, and are rarely analyzed through the lens of quality-cost trade-offs. We introduce \textbf{\thinkbooster}, a unified framework for seamless test-time compute scaling of LLM reasoning, which consists of (i) a modular Python library implementing %nine 
state-of-the-art TTC scaling strategy and scorer families, 
%including process reward models (PRMs), uncertainty-based methods, LLM-as-a-critic, and internal-state probing, 
(ii) a benchmark that jointly evaluates performance and computational efficiency, and (iii) a deployable OpenAI-compatible proxy service that enables drop-in integration of adaptive reasoning into real-world applications. We further provide a demo visual debugger for inspecting the reasoning trajectories, intermediate selection decisions, and alternative reasoning paths.\footnote{\url{http://demo-thinkbooster.nlpresearch.group}}\textsuperscript{,}\footnote{\url{http://video-thinkbooster.nlpresearch.group}} Empirical results on mathematical and coding tasks reveal the performance-compute trade-offs of TTC scaling strategies and scoring methods and demonstrate that \thinkbooster provides practical gains in real-world tasks. The code is available online under an MIT license.\footnote{\url{http://thinkbooster.nlpresearch.group}}
%\thinkbooster makes TTC scaling easy to use a
%demonstrate that principled, uncertainty-aware escalation strategies achieve superior quality–cost trade-offs compared to static scaling baselines. 
%\thinkbooster operationalizes test-time compute scaling as a controllable systems knob, enabling reliable and compute-aware reasoning in both research and production settings.
\end{abstract}

\section{Introduction}
% Provide an overview of the problem of test-time compute scaling for reasoning and introduce the ThinkBooster framework.

\begin{figure}
    \centering
    \includegraphics[width=0.8\linewidth,trim=0mm 0mm 0mm 0mm,clip]{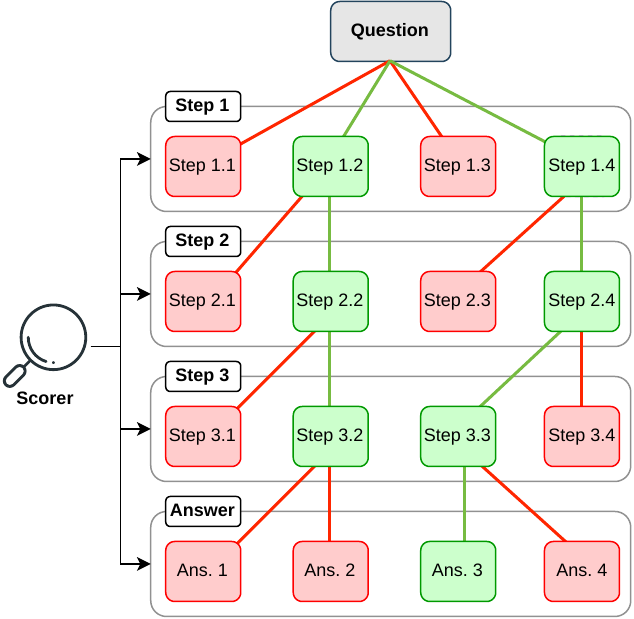}
    \caption{Illustration of the common reasoning test-time compute scaling strategy -- beam search (tree of thought with breadth first search).}
    \label{fig:totimage}
    % https://drive.google.com/file/d/1bW3KAK5JlvRkf2vgzqS6bG1bulf-ABEW/view?usp=sharing
\end{figure}

\begin{figure*}
    \centering
    \includegraphics[width=0.75\linewidth,trim=5mm 4mm 5mm 4mm,clip]{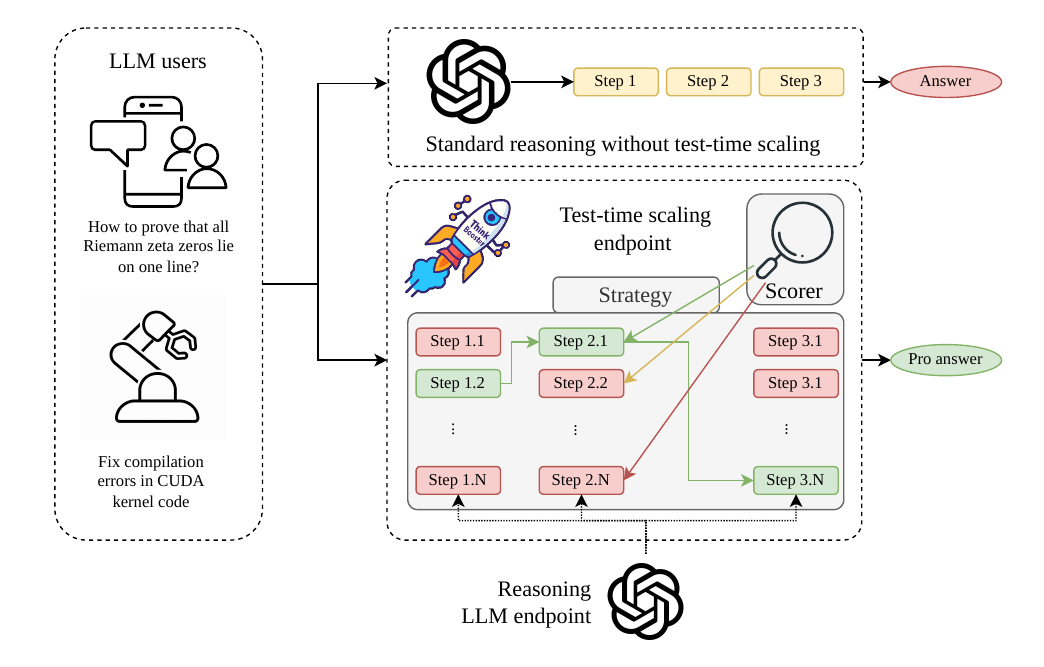}
    \caption{An illustration of the \thinkbooster endpoint gateway for test-time compute scaling.}
    \label{fig:proxy_service}
    % https://drive.google.com/file/d/1bUI90FmeBD5MmckxlzcV3GHwCNV-uRIX/view?usp=sharing
\end{figure*}

Chain-of-thought (CoT) prompting \cite{10.5555/3600270.3602070,kojima2022large} and, more recently, fine-tuning Large Language Models (LLMs) to natively produce intermediate reasoning before a final answer \cite{lightman2023let,deepseek2025r1} have unlocked powerful capabilities in LLMs to solve complex tasks in mathematics, programming, and even scientific research \cite{10.5555/3600270.3602070,chen2021evaluating,lu2024aiscientist}.

It has also been shown that increasing compute at test time, such as by generating longer reasoning chains or sampling multiple solutions and selecting the best ones, can significantly improve performance on challenging problems. This approach is known as \emph{test-time compute (TTC) scaling} \cite{ICLR2025_1b623663,muennighoff2025s1}.
%It was found to be extremely useful in situations where an LLM is already state-of-the-art yet still fails to solve a challenging problem in a single pass. TTC scaling also serves as the primary approach for improving performance  when the underlying LLM is fixed and cannot be further fine-tuned. 
It is especially effective when a state-of-the-art LLM fails to solve a challenging problem in a single pass and cannot be further fine-tuned. Moreover, recent work shows that TTC scaling can provide a better performance to efficiency trade-off than simply increasing model size \cite{ICLR2025_1b623663}.

Common TTC scaling strategies include best-of-$N$ (BoN: \citet{cobbe2021training,ICLR2025_1b623663}), tree-of-thought (ToT: \citet{yao2023tree}, see \Cref{fig:totimage}), and self-consistency (also known as majority voting: \citet{DBLP:conf/iclr/0002WSLCNCZ23}).
Recently, there have been a number of proposed dynamic TTC scaling strategies that adapt the compute spent based on confidence or uncertainty estimated at the level of individual reasoning steps \cite{zhang-etal-2025-entropy,fu2025deepthinkconfidence,yan2025murmomentumuncertaintyguided}.

Another line of research develops step- and trajectory-level scorers to select the most promising reasoning path from multiple candidates.
%Another line of research is the development of reasoning step and trajectory scorers that can be used to select the most prominent reasoning paths from a set of candidates. 
Common approaches include verification via process reward models (PRMs) \cite{uesato2022solving,lightman2023let,li-etal-2023-making,luo2024improve,zhang-etal-2025-lessons} and self-verification via the same LLM \cite{xie2023self,weng-etal-2023-large}. Recent work has also proposed unsupervised and supervised uncertainty or confidence scores to assess the reliability of the reasoning steps \cite{kadavath2022language,zhu2025uncertainty,ni2025efficient}. 

While TTC scaling has seen rapid advancements, the literature on this topic remains highly fragmented. Methods are typically evaluated under different experimental protocols, model configurations (e.g., structured CoT vs.\ native unstructured thinking), and compute budgets, which makes direct comparison difficult. Moreover, many studies focus primarily on accuracy gains while overlooking the associated computational costs, latency, and efficiency trade-offs. 
As a result, it is difficult to identify which methods truly offer the best performance-compute trade-off. 
%thereby hindering systematic progress in the field. 
Moreover, most published research typically presents implementations of only a single proposed method, along with a limited set of baselines. Finally, the released code often serves merely as a proof of concept, exhibiting efficiency limitations and lacking support for practical deployment. The lack of standardized, reliable, and efficient implementations creates an additional challenge for practitioners deploying TTC scaling in real-world applications.
In this work, we bridge these gaps by presenting \thinkbooster, a unified framework for seamless test-time compute scaling of LLM reasoning. \thinkbooster targets NLP researchers studying reasoning and test-time compute scaling, as well as practitioners deploying LLM-based applications. Our goals are two-fold: (i) to provide a unified framework for principled benchmarking and research on TTC scaling and reasoning in general, and (ii) to provide a practical developer-oriented integration layer that supports easy deployment of TTC scaling in real-world applications through a unified API and clear abstractions. The framework combines a modular Python library, a benchmark, and a deployable TTC scaling endpoint gateway (see \Cref{fig:proxy_service}) that provides TTC scaling as a service and can be used as a drop-in replacement for an OpenAI-compatible LLM endpoint.  Acting as a transparent proxy to the underlying LLM, the \thinkbooster endpoint gateway seamlessly applies test-time compute scaling on top of the underlying LLM generations, thus enhancing the quality of the reasoning trajectories and the final answers without the need for any modifications to the existing application logic. It effectively equips the underlying LLM with a ``Pro'' reasoning mode. By improving the quality of the final LLM answers, \thinkbooster directly enhances the reliability and effectiveness of downstream applications built on top of LLMs, including AI agents. Finally, we provide a demo visual debugger for reasoning trajectories that can be used to inspect the intermediate steps, selection decisions, and alternative paths.

\begin{table*}
    \centering
    \footnotesize
    \resizebox{\textwidth}{!}{%
    \begin{tabular}{llllc}
        \toprule
        \textbf{Method} & \textbf{Key idea}  & \multirowcell{\textbf{Offline} / \\ \textbf{Online}} & \multirowcell{\textbf{Level of} \\ \textbf{access to} \\ \textbf{LLM}} & \multirowcell{\textbf{Needs} \\ \textbf{prefill}} \\
        \midrule
         Best of N \cite{cobbe2021training} & Sample $N$ solutions and select the best one & Offline & Black-box & No \\
         Majority voting \cite{DBLP:conf/iclr/0002WSLCNCZ23} & Sample N solutions and select answer by majority vote & Offline & Black-box & No \\
         Beam search (ToT) \cite{yao2023tree,xie2023self} & Explore tree of reasoning paths and select best & Online & Black-box & Yes \\
         Extended thinking \cite{muennighoff2025s1} & Control reasoning budget to force longer CoT & Online & Black-box & Yes \\
         Dynamic exploration, MUR \cite{zhang-etal-2025-entropy,yan2025murmomentumuncertaintyguided} & Only allocate more compute on uncertain steps  & Online & White-box &  Yes\\
         DeepConf online \cite{fu2025deepthinkconfidence} & Steer generation toward high-confidence tokens & Online & White-box & Yes \\
         DeepConf offline \cite{fu2025deepthinkconfidence} & Rerank candidate solutions by model confidence scores & Offline & White-box & No \\
         Phi-decoding \cite{xu2025phidecodingadaptiveforesightsampling} & Foresight sampling and adaptive pruning based on an uncertainty signal & Online & White-box & Yes \\
         Uncertainty CoT \cite{zhu2025uncertainty} & Generate multiple trajectories when uncertain & Online & White-box & Yes \\
         \bottomrule
    \end{tabular}
    }
    \caption{Test-time compute scaling strategies implemented in \thinkbooster.}
    \label{tab:tts_strategies}
\end{table*}

\begin{table*}
    \centering
    \scriptsize
    \resizebox{\textwidth}{!}{
    \begin{tabular}{lll}
    \toprule
    \textbf{Method}     &  \textbf{Key idea} & \multirowcell{\textbf{Level of} \\ \textbf{access to} \\ \textbf{LLM}} \\
    \midrule
        Process reward models \cite{lightman2023let} & Train a separate critic LLM to score reasoning steps and trajectories & Black-box \\
        Self-verification \cite{yao2023tree,weng-etal-2023-large} & Ask the same LLM to evaluate the reasoning step / trajectory & Black-box\\
        Uncertainty and confidence \cite{zhang-etal-2025-entropy,fu2025deepthinkconfidence} & Confident step options are higher priority & White-box \\
        ReProbes \cite{ni2025efficient,shelmanov-etal-2025-head} & A supervised regressor on top of the LLM internal states & White-box\\
    \bottomrule
    \end{tabular}
    }
    \caption{Reasoning step and trajectory scorers implemented in \thinkbooster.}
    \label{tab:tts_scorers}
\end{table*}

\thinkbooster lowers the barrier for adopting TTC scaling for both researchers and practitioners.
The contributions of this work are as follows:
\begin{itemize}
% [itemsep=0pt,topsep=0pt] % Better do NOT mess with formatting; it can lead to a desk-rejection; has happened to me a couple of times. You can save 1-2 lines in so many other ways. Not worth the risk in my view.
    % \item The library of Python implementations of state-of-the-art TTC scaling strategies and scorers with a unified program API.
    \item A Python library that implements state-of-the-art TTC scaling strategies and scorers behind a unified and consistent programming API.
    \item A practical TTC scaling endpoint %gateway 
    with an OpenAI-compatible remote API that can be used as a drop-in replacement for the original LLM endpoint in various applications, including AI agents and assistants. We show that \thinkbooster can improve downstream performance in real-world tasks, using CUDA kernel optimization as an example.
    \item A benchmark for conducting research in reasoning and test-time compute scaling with both performance and compute metrics. We also conduct a pilot study of implemented strategies and scorers and provide insights into their performance-efficiency tradeoffs.
    \item Finally, we create a demo visual debugger for LLM reasoning trajectories during TTC scaling, which enables interactive inspection of the intermediate reasoning steps and facilitates the systematic analysis of LLM's errors.  
\end{itemize}

\section{Core Library}
\label{sec:library}

\thinkbooster, at its core, is a lightweight, modular, and extendable Python library that implements the key components for TTC scaling. These include scaling strategies, scorers, reasoning generators, and reasoning step boundary detectors.
% TODO: add reference to library structure figure (fig:library) when appendix space allows

\ecoparagraph{Scaling strategies} define the high-level scaling algorithm for test-time compute scaling, but typically do not prescribe low-level details, such as how exactly the reasoning steps or trajectories are scored. Our library includes implementations of nine state-of-the-art algorithms (see \Cref{tab:tts_strategies}), which cover more than twenty recent publications on TTC scaling and reasoning. 
%Typically, the strategies in the library define the high-level scaling algorithm and do not prescribe low-level implementation details, such as how the reasoning steps or trajectories are scored. 
Some strategies operate offline, meaning that they enable scoring trajectories after the final solution is obtained, while others perform scoring online as part of the reasoning process. They also differ by their level of access to LLM outputs and internal states. \emph{White-box} strategies require access to logits or internal states, which limits their applicability in certain LLM deployment settings. 

\emph{Black-box} strategies do not need anything except for the generated tokens. Finally, some strategies require the \emph{Prefill} option to be enabled in the provider's API -- it allows to populate the textual prefix with previously generated reasoning steps, so that the LLM does not start reasoning from scratch.

\ecoparagraph{Scorers} specify the way in which individual reasoning steps or entire trajectories are assessed. We implement four major approaches (see \Cref{tab:tts_scorers}): (1) PRMs \cite{lightman2023let}, (2) uncertainty scores and confidence scores based on the LM-Polygraph library \cite{fadeeva2023lm,vashurin-etal-2025-benchmarking}, (3) LLM-as-a-judge assessment (including self-assessment: \citet{yao2023tree}), and (4) ReProbes \cite{ni2025efficient}. The scorers can differ in their required level of access to the internal states or outputs of the underlying LLM:
white-box vs. black-box.

\ecoparagraph{Reasoning generators} are wrappers around LLM deployments. The library supports LLMs deployed locally via the Hugging Face Transformer library and vLLM \cite{kwon2023efficient}. An LLM could also be deployed as a service via vLLM, OpenRouter, ChatGPT, or by any other provider % such as Gemini or DeepSeek 
with an OpenAI compatible API. Usually, the most flexible white-box access can be obtained through the Transformers and vLLM API. The majority of providers expose only generated tokens, but some give access to the token log-probabilities (e.g., gpt-4o in OpenAI, some LLMs in OpenRouter, and the DeepSeek API). The prefill option is also available for vLLM deployments and some providers, such as DeepSeek and Anthropic.

\ecoparagraph{Reasoning step extractors} enable the decomposition of the reasoning trajectory into atomic segments and allow controlled pauses in generation for processing individual steps. For non-thinking LLMs, one can specify a system prompt to facilitate CoT in a certain format available for parsing. For large reasoning language models (LRLMs) with a native thinking mode, such as DeepSeek R1 or Qwen 3, it is not possible to control the format of the thoughts inside \verb|<think>...</think>| tags, which complicates reliable step extraction.
We support the extraction of reasoning steps from both structured generation facilitated by the system prompt and unstructured thinking in LRLMs. 

% The library allows combining various TTC scaling strategies, scorers, LLMs, and reasoning step extractors to obtain a configuration that is suitable for the needs of developers and researchers. The key aspect for the success of TTC scaling in practical use cases is the efficiency and reliability of implementation. To make our implementation practical, we 
% use the maximum of the abilities of provider APIs

The library enables flexible combinations of scaling strategies, scoring mechanisms, LLM backends, and reasoning step extractors, tailored to specific needs. 
%Previous related work has focused on non-thinking LLMs, did not allow combinations of different strategies and scorers, and primarily focused on evaluating reasoning chains \cite{hao2024llm}.
A key requirement for the practical adoption of TTC scaling is the efficiency and reliability of its implementation. We address this by engineering optimized components that fully exploit provider APIs. For example, we implement custom wrappers around vLLM that expose the hidden states during inference, thus enabling efficient integration of internal-state-based scoring without requiring modifications to the underlying model.

\begin{figure}[t]
\centering

{\scriptsize\begin{verbatim}
from openai import OpenAI

client = OpenAI(
    base_url="<THINKBOOSTER_ENDPOINT>/v1/beam_search/prm",
    api_key="<YOUR_API_KEY>",
)
response = client.chat.completions.create(
    model="Qwen/Qwen3-30B-A3B",
    messages=[{"role": "user", "content":
        "Find the number of ordered pairs (x, y) of "
        "positive integers satisfying x + 2y = 2xy."}],
    extra_body={ # Optional parameters
        "model_base_url": "https://openrouter.ai/api/v1",
        "max_tokens": 8192, "tts_beam_size": 4,
    },
)
print(response.choices[0].message.content)
# reasoning path with answer
\end{verbatim}}

\caption{Accessing the \thinkbooster endpoint gateway through the OpenAI Python SDK. Parameter \texttt{base\_url} specifies the \thinkbooster endpoint URL, which encodes scaling strategy and scorer. 
%Additional parameters can be passed via \texttt{extra\_body}.
}
\label{fig:code_example}
\end{figure}

\begin{figure}[t]
    \centering
    \begin{subfigure}{\linewidth}
        \centering
        \includegraphics[trim=75pt 490pt 75pt 123pt, clip, width=\linewidth]{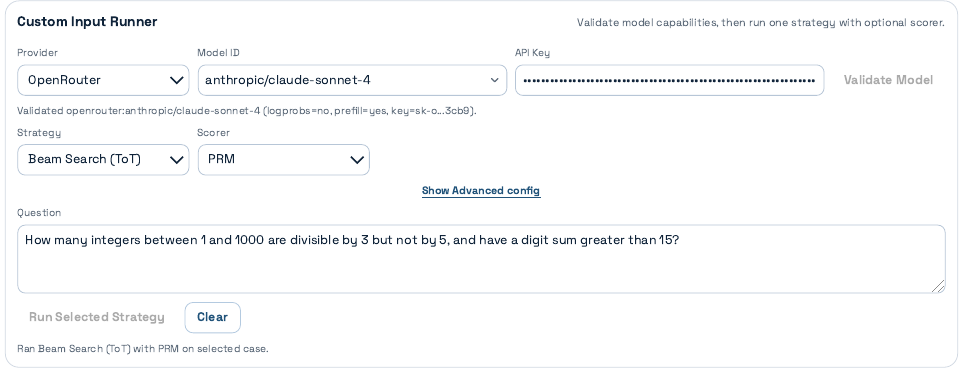}
        \caption{Customization of provider, model, strategy, and scorer.}
        \label{fig:debugger-parameters}
    \end{subfigure}

    \vspace{0.6em}

    \begin{subfigure}{\linewidth}
        \centering
        \includegraphics[trim=75pt 105pt 75pt 305pt, clip, width=\linewidth]{main-1-new.pdf}
        \caption{Reasoning timeline and step inspector with per-candidate score breakdowns.}
        \label{fig:debugger-timeline}
    \end{subfigure}

    \caption{Visual debugger for reasoning.}
    % See details in Appendix~\ref{app:demo-ui}.}
    \label{fig:debugger}
\end{figure}

\begin{figure}[t]
    \ContinuedFloat
    \setcounter{subfigure}{2}% next subfigure = (c)
    \centering
    \begin{subfigure}{\linewidth}
        \centering
        \includegraphics[trim=120pt 360pt 120pt 40pt, clip, width=0.55\linewidth]{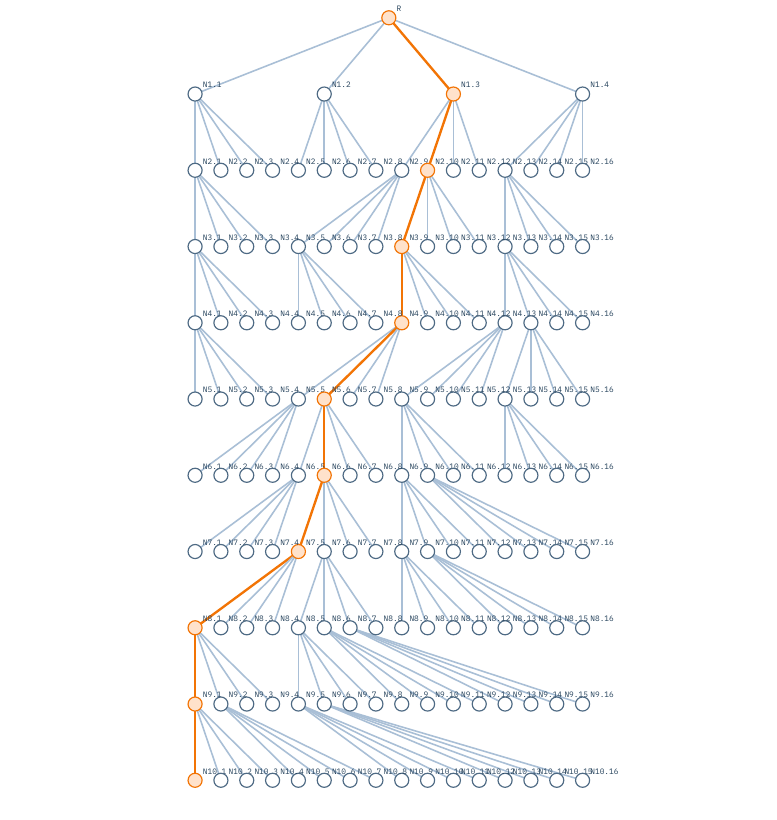}
        \caption{Trajectory tree: orange path = selected run, siblings = pruned candidates.}
        \label{fig:debugger-tree}
    \end{subfigure}
\end{figure}

\section{Endpoint Gateway for TTC Scaling}

For developers and end-users, we built a dedicated endpoint gateway that provides TTC scaling as a service. It enhances model outputs without requiring any modifications to existing application logic (see \Cref{fig:proxy_service}), operating as a proxy in front of the underlying LLM, applying TTC scaling before returning the response. 
The user just needs to replace the original LLM endpoint URL with the URL of the \thinkbooster endpoint (see \Cref{fig:code_example}).

The \thinkbooster endpoint supports the configuration of TTC scaling parameters directly through a URL or an API, allowing users to control compute budgets, reasoning strategies, and scorers. %This design enables seamless integration of \thinkbooster into any application that already relies on a standard LLM service.
As the endpoint preserves the standard LLM interface, it can be integrated into downstream systems such as AI agents, code assistants, or enterprise copilots without any refactoring. Endpoint URLs can be easily configured in the code or by specifying the environment variables. \thinkbooster allows developers to enable a ``Pro reasoning mode'' for any OpenAI-compatible LLM deployment, improving quality of final answers while maintaining explicit control over computational costs. 

\section{Demo Visual Debugger for Reasoning}
\label{sec:demo}

To support the analysis of TTC scaling algorithms and scorers, we provide an interactive Visual Debugger for Reasoning. The tool allows users to inspect intermediate trajectories and reasoning steps.
%generated by various TTC scaling strategies, including sampled candidates, tree expansions, and verifier based reranking.
At each reasoning step, it exposes scores (e.g.\ uncertainty / confidence / PRM scores), making it possible to trace decisions, alternative reasoning paths, and understand why a particular trajectory is selected (see \Cref{fig:debugger}).

% Compared to prior work that visualizes reasoning traces \cite{li-etal-2025-reasongraph}, our debugger is natively integrated with a modular TTC scaling framework that supports diverse strategies and scorers, enabling side-by-side trajectory comparison.
%Screenshots of the Visual Debugger UI are provided in Appendix~\ref{app:demo-ui}.

% The debugger can be used both as a research tool and as an educational interface. Researchers can compare strategies under identical prompts and compute budgets, analyze failure cases, and diagnose suboptimal escalation thresholds. 
%Practitioners can observe how additional compute is allocated across reasoning steps and better understand the quality–co

\section{Related Work}
\label{sec:related}

%Missing SkyThought
% they can be plugged in as a black-box trajectory scorer (cf.\ self-verification in \Cref{tab:tts_scorers}) or as a refinement stage in our pipeline.
% % , and are target for future releases.

%\ecoparagraph{
Comparison with other open-source TTC scaling frameworks is presented in \Cref{tab:ttc_frameworks}. 
%Several open-source frameworks implement TTC scaling, but each targets a limited part of \thinkbooster's features. 
% \Cref{tab:ttc_frameworks} provides a feature-level comparison against six representative systems. 
OptiLLM~\cite{optillm} is the closest in scope -- an OpenAI-compatible proxy that implements several TTC scaling techniques. However, it lacks PRM scorers, FLOPs-level compute accounting, and a visual reasoning debugger. 

LLM Reasoners~\cite{hao2024llm} provides a modular library with implementations of search algorithms, reward functions, methods for reasoning correction, and a reasoning tree visualizer. However, it has no OpenAI-compatible endpoint, no native vLLM backend, and no joint performance-compute benchmark. 
% a clean WorldModel\,+\,Search abstraction with a hosted tree visualizer, but it has no OpenAI-compatible endpoint, no native vLLM backend, and no joint performance-compute benchmark. 
OpenR~\cite{wang2024openr} implements two TTC scaling strategies: beam search and best-of-$N$ with PRM scorers and is coupled with the FastChat inference framework (vLLM analog). 
%The main feature of OpenR is data augmentation and RL trianing rather than TTC scaling.
%combines step-level PRM-guided search with PRM training but is tightly coupled to FastChat orchestration and math tasks. 
search-and-learn~\cite{beeching2024scalingtesttimecompute} is a limited set of TTC scaling methods for reproducing the experiments from \cite{ICLR2025_1b623663} on the MATH-500 dataset. 
% vLLM-only reproduction of limited number of TTC scaling methods from \cite{ICLR2025_1b623663} for the MATH-500 dataset.
%reproducibility recipe set for verifier-guided search on MATH-500. 
TreeQuest~\cite{misaki2025adaptivebranchingmontecarlotree} implements adaptive branching MCTS over user-defined state generators with visualization utilities. 
%TextGrad~\cite{yuksekgonul2024textgrad} optimizes prompts via textual gradients and does not expose search/scoring over reasoning trajectories.
ReasonGraph~\cite{li-etal-2025-reasongraph} provides basic implementations of TTC scaling techniques, but its primary focus is on visualizing reasoning paths. We deliberately omit self-correction and evolutionary search methods \cite{Romera-Paredes2024,kumar2025trainingselfcorrect,liu2025trustverify,10.5555/3666122.3668141}, as they present an orthogonal line of work, where the TTC scaling considered in this work could be plugged-in for better performance.

\thinkbooster stands out as a uniquely feature-complete framework. It provides a comprehensive suite of TTC scaling strategies and scorers, seamless compatibility with efficient LLM-serving frameworks like vLLM, and a benchmark comprising nine bundled math, coding, and scientific datasets with joint TFLOPs-and-tokens compute accounting. Additionally, the framework equips practitioners with an OpenAI-compatible API endpoint and an interactive visual reasoning debugger.

% \ecoparagraph{Self-correction and evolutionary search} is a parallel line of work, in which the model is trained or prompted to revise its own solutions \cite{Romera-Paredes2024,kumar2025trainingselfcorrect,liu2025trustverify}.
% % before committing to a final answer. 
% %These methods are orthogonal to those in \thinkbooster. 
% TTC scaling methods discussed in this work could be integrated into such pipelines to further improve their performance.

\begin{table*}[t]
\centering
\footnotesize
\setlength{\tabcolsep}{4pt}
\resizebox{\textwidth}{!}{%
\begin{tabular}{lccccccc}
\toprule
\textbf{Feature} & \textbf{\thinkbooster} & \textbf{OptiLLM} & \textbf{LLM Reasoners} & \textbf{OpenR} & \textbf{S\&L} & \textbf{TreeQuest} & \textbf{ReasonGraph} \\
\midrule
\textbf{A.} Strategy taxonomy breadth & \checkmark\ (9) & $\circ$ (7) & $\circ$ (5) & $\circ$ (5) & $\times$ (3) & $\times$ (1) & $\times$ (0) \\
\textbf{B.} Scorer family breadth & \checkmark\ (4) & $\circ$ (2) & $\circ$ (2) & $\times$ (1) & $\times$ (1) & $\circ$ (1) & $\times$ (0) \\
\textbf{C.} Supports TTC methods up to year & \textbf{2026} & 2026 & 2024 & 2024 & 2024 & 2026 & 2023 \\
\textbf{D.} Supports uncertainty-based scorers & \checkmark & $\times$ & $\times$ & $\times$ & $\times$ & $\times$ & $\times$ \\
\textbf{E.} Joint perf--compute benchmarks (TFLOPs + tokens) & \checkmark & $\circ$ & $\circ$ & $\circ$ & $\circ$ & $\times$ & $\times$ \\
\textbf{F1.} Bundled math benchmarks (count) & \textbf{5} & 0 & 0 & 2 & 1 & 0 & 0 \\
\textbf{F2.} Bundled coding benchmarks (count) & \textbf{3} & 0 & 0 & 0 & 0 & 0 & 0 \\
\textbf{G.} Backends supported & vLLM+HF+API & API+HF+MLX & HF+SGLang+API & vLLM+HF & vLLM & agnostic & API \\
\textbf{H.} OpenAI-compatible REST gateway & \checkmark & \checkmark & $\times$ & $\times$ & $\times$ & $\times$ & $\times$ \\
\textbf{I.} Visual debugger & \checkmark & $\times$ & $\circ$ & $\circ$ & $\times$ & $\times$ & \checkmark \\
\textbf{J.} Modular architecture & \checkmark & $\circ$ & $\circ$ & $\circ$ & $\times$ & $\circ$ & $\circ$ \\
\bottomrule
\end{tabular}
}
\caption{\label{tab:ttc_frameworks}
Comparison of \thinkbooster with other TTC scaling libraries and reasoning-analysis frameworks: OptiLLM~\cite{optillm}, LLM~Reasoners~\cite{hao2024llm}, OpenR~\cite{wang2024openr}, search-and-learn (S\&L)~\cite{beeching2024scalingtesttimecompute}, TreeQuest~\cite{misaki2025adaptivebranchingmontecarlotree}, %TextGrad~\cite{yuksekgonul2024textgrad}, 
and ReasonGraph~\cite{li-etal-2025-reasongraph}.
Symbols: \checkmark\ supported, $\circ$ partial / limited, $\times$ not supported. ``API'' denotes any OpenAI-compatible HTTP endpoint (OpenAI, Anthropic, Gemini, OpenRouter, etc.). Strategy counts include genuine search/scoring/decoding-intervention methods only (prompt-engineering scaffolds excluded). Bundled benchmark counts include only datasets with prewired prompts, answer extraction, and judging.
% Counts in rows~A--B and rows~F1--F2 reflect the framework's first-class abstractions/datasets; see Appendix~\ref{app:framework_comparison} for per-cell rationale.
}
\end{table*}

\section{Experiments}

\subsection{Experimental Setup}

% We developed a benchmarking script for the systematic evaluation of scaling strategies and scorers implemented in \thinkbooster, as well as custom user-defined methods, across diverse datasets and LLMs. 

We introduce a benchmarking tool designed for the systematic evaluation of \thinkbooster's built-in scaling strategies and scorers, as well as custom user-defined methods, across a diverse range of datasets and LLMs.
Using this tool, we conducted a pilot study on the performance-efficiency trade-off of popular methods.

\vspace{0.1cm}
\ecoparagraph{LLMs for reasoning.}
We experimented with three state-of-the-art LLMs: Qwen2.5-Math-7B-Instruct~\cite{yang2024qwen2} without thinking mode, Qwen3-8B in native thinking mode~\cite{yang2025qwen3}, and a large GPT-OSS-120B~\cite{openai2025gptoss120bgptoss20bmodel}. The selected models span distinct categories: non-thinking, small-thinking, and large-thinking LLMs, achieving state-of-the-art performance within their respective groups.

%were chosen for their high performance in reasoning tasks for their size and because they represent different categories
%It has been chosen for its mathematical abilities. The performance of this model is measured across eight strategies, six scorers and four mathematical datasets. We compare the metrics with the baseline Chain-of-Thought (Raw CoT) reasoning where no strategies are applied.

% Qwen3~\cite{yang2025qwen3} implements the native thinking mode which is used to evaluate reasoning strategies. Since the thinking process in Qwen3 is generated as a whole text (as opposed to separate steps in some other models), we use either linguistic markers (for mathematical datasets) or new line separators (for coding datasets). TODO: add the whole list into appendix
% math for qwen2.5: math500, olympiad bench, minerva math, gaokao23en\\
% math for qwen3 thinking:aime24-25\\

\vspace{0.1cm}
\ecoparagraph{Datasets.} We select challenging datasets from three categories: \emph{mathematics}: MATH-500~\cite{math500}, OlympiadBench~\cite{he2024olympiadbench}, GaoKao23EN~\cite{zhang2023evaluating}, AIME-2024, and AIME-2025~\cite{aime_problems_solutions}; \emph{scientific QA}: GPQA-Diamond~\cite{rein2023gpqa}; and \emph{coding}: HumanEval+, MBPP+~\cite{liu2023your}, and KernelBench~\cite{ouyang2025kernelbench}.

% The datasets in the required format are available in our repository.\footnote{\url{https://huggingface.co/test-time-compute}} Some benchmarks appear to be saturated for particular LLMs. Therefore, we report results for a given model-dataset pair only when the dataset remains sufficiently challenging for that model.
% The model-dataset mapping, dataset statistics, and further details are in \Cref{app:exp_details}.

We provide the appropriately formatted datasets via our repository.\footnote{\url{https://huggingface.co/test-time-compute}} Because certain benchmarks have reached saturation for specific LLMs, we report results only for model-dataset pairs where the task remains sufficiently challenging. Detailed dataset statistics and exact model-dataset mappings can be found in \Cref{app:exp_details}.

%, as these low-level primitives serve as the building blocks for higher-level functions. 
%in KernelBench, as these low-level primitives serve as the building blocks for higher-level functions.
% Maybe the below sentence is not needed.
%Overall, the complexity of the task and the real world impact are an excellent testing ground for enhanced reasoning abilities of the model.

% For \textbf{coding} datasets, we use HumanEval+ and MBPP+ \cite{liu2023your}, which are extensions of the original HumanEval \cite{chen2021evaluating} and MBPP \cite{austin2021program} benchmarks. Both enhanced datasets were introduced by Liu et al. as part of the EvalPlus framework, which substantially expands the number of test cases per problem — up to 80 times more than the originals. 

\begin{figure*}[t]
    \centering
    \begin{subfigure}[t]{0.49\textwidth}
        \centering
        \includegraphics[width=\linewidth]{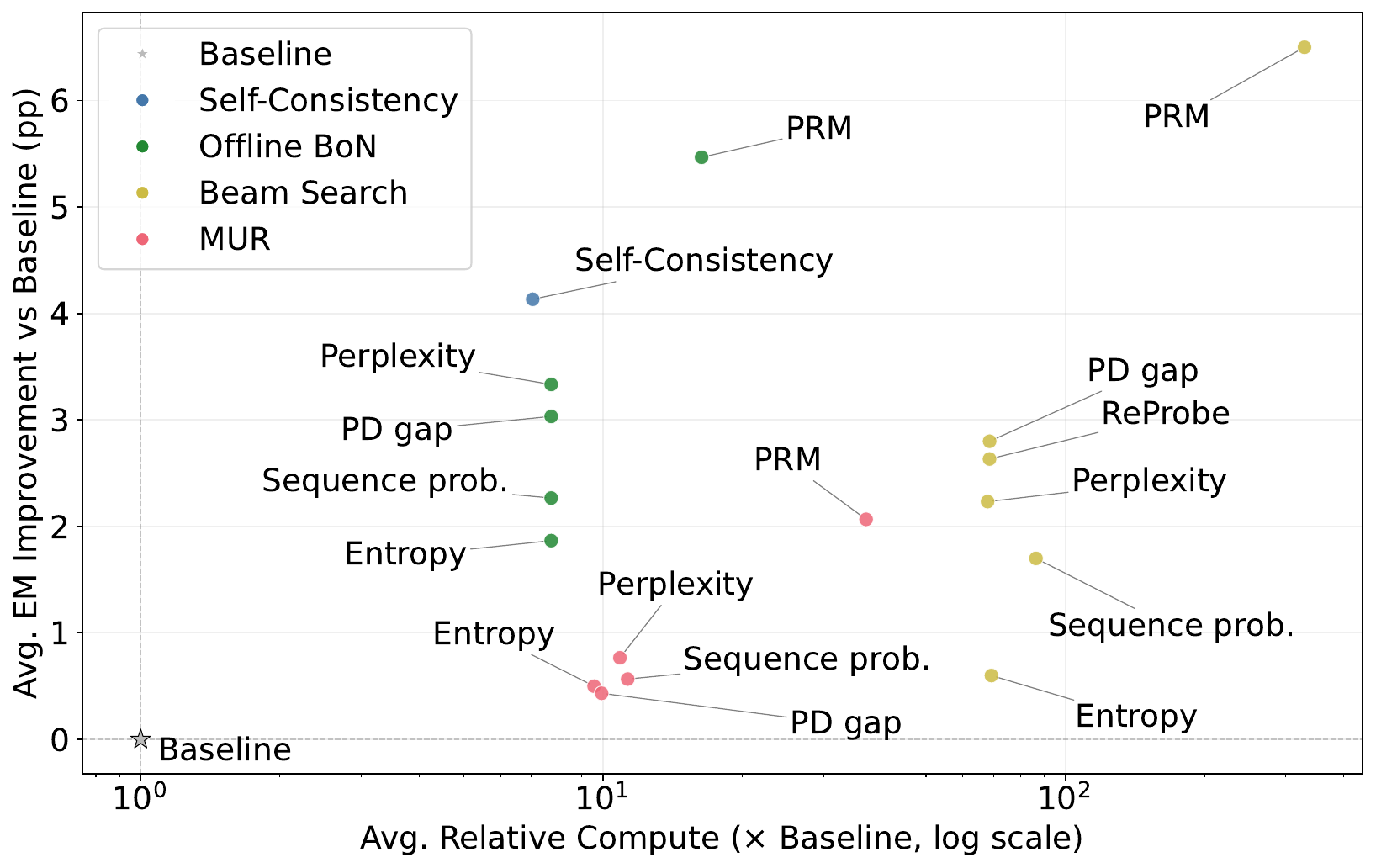}
        \caption{Qwen2.5-Math-7B (aggregate over MATH-500, OlympiadBench, Gaokao~2023~EN)}
        \label{fig:qwen25-ratio}
    \end{subfigure}
    \hfill
    \begin{subfigure}[t]{0.49\textwidth}
        \centering
        \includegraphics[width=\linewidth]{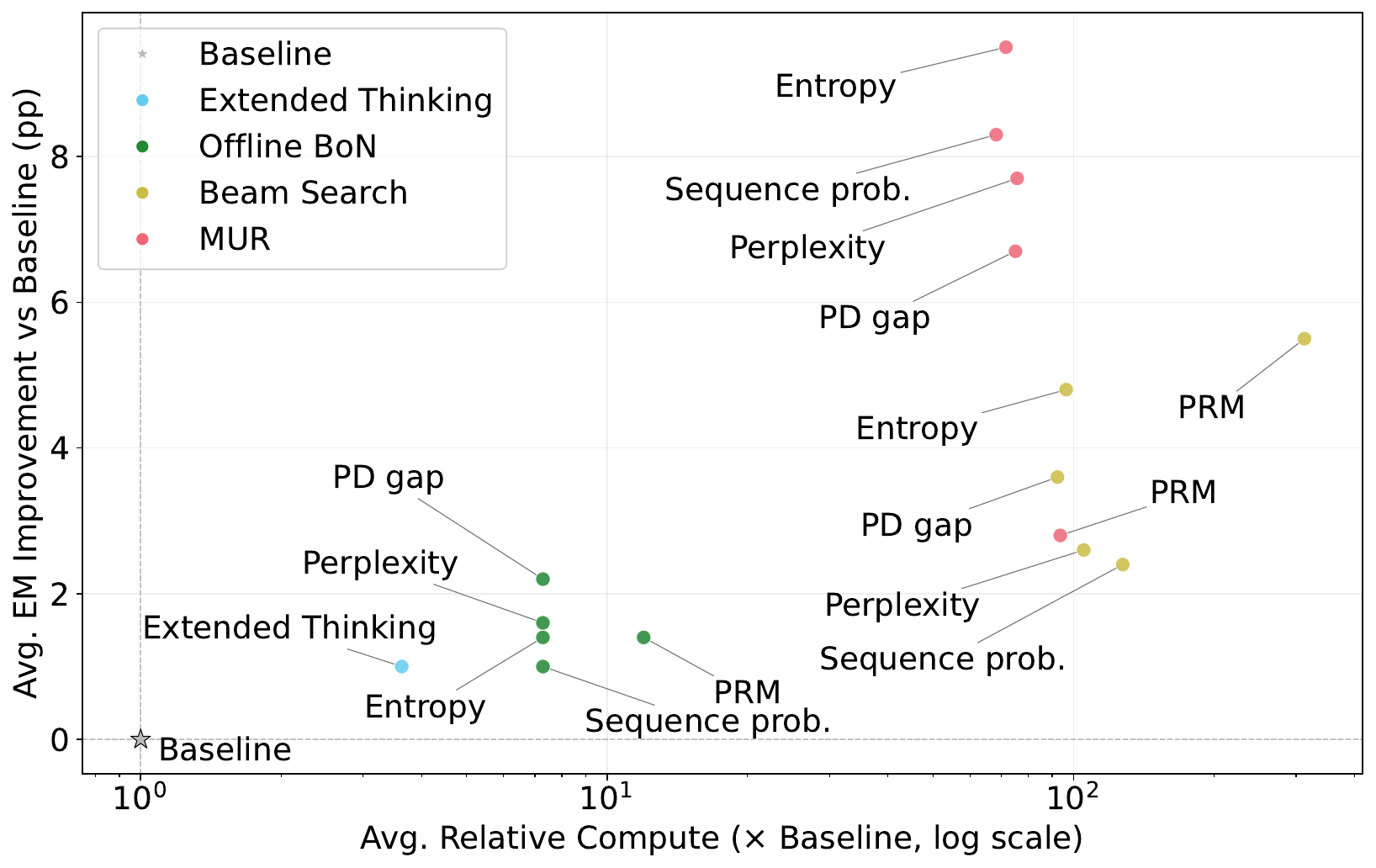}
        \caption{Qwen3-8B (HumanEval-Plus)}
        \label{fig:qwen3-humaneval-ratio}
    \end{subfigure}
    \caption{Accuracy improvement vs.\ compute ratio for different TTC scaling methods. Each point represents a strategy-scorer combination; the $x$-axis shows the compute ratio relative to the baseline (log scale). Full per-dataset results are in Appendix~Tables~\ref{tab:qwen25-full-results}~and~\ref{tab:qwen3-full-results}.}
    \label{fig:accuracy-vs-compute}
\end{figure*}

\begin{table*}[t]
\centering
\footnotesize
\resizebox{0.8\textwidth}{!}{%
\begin{tabular}{l|c|c|c|ccc}
\toprule
\multirow{2}{*}{\textbf{Strategies}} &
\multicolumn{1}{c|}{\textbf{GPQA-Diamond}} &
\multicolumn{1}{c|}{\textbf{HumanEval+}} &
\multicolumn{1}{c|}{\textbf{MBPP+}} &
\multicolumn{3}{c}{\textbf{KernelBench}} \\
\cline{2-7}
% \cmidrule(lr){2-7}%\cmidrule(lr){3-3}\cmidrule(lr){4-4}\cmidrule(lr){5-7}
&
\textbf{EM} &
% \textbf{LLM-as-a-Judge} &
\textbf{pass@1} &
\textbf{pass@1} &
\textbf{Syntax} &
\textbf{Compilation} &
\textbf{Correctness} \\
\midrule
Raw CoT &
70.3 &
% 71.7 &
83.5 &
76.0 &
82.0 &
\textbf{65.0} &
26.0 \\

Offline BoN &
72.2 &
% 71.9 &
85.4 &
\textbf{78.8} &
\textbf{87.0} &
64.0 &
\textbf{30.0} \\

Beam Search &
\textbf{73.2} &
% -- &
\textbf{87.8} &
78.4 &
-- & -- & -- \\
\bottomrule
\end{tabular}%
}
\caption{\label{tab:gpt_oss}
Evaluation results for \texttt{gpt-oss-120b} using a PRM-based scorer across QA (GPQA-Diamond), coding (HumanEval+, MBPP+), and CUDA kernel generation (KernelBench). The best results are highlighted in \textbf{bold}.}
\end{table*}

\vspace{0.1cm}
\ecoparagraph{Performance metrics} and the result parsing procedures are dataset-specific. We take their implementations from the corresponding papers for consistency and more direct comparison. For mathematical and scientific datasets, usually, 
%(MATH500, Olympiad Bench, %Minerva Math, 
%GaoKao23En, AIME), 
we can determine whether the LLM response matches the gold answer exactly (exact match = EM). 
%For GPQA-Diamond, we also report the exact match metric.
% Performance + compute metrics
% , we rely on the functions reported in the original model papers. Thus, we use the implementation for the exact match function from the Qwen2.5 technical report~\cite{qwen2024qwen2}.
% We use appropriate metrics for each dataset. 
% Since all math datasets expect some form of a math expression (MATH500, Olympiad Bench, Minerva Math, GaoKao23En) or an integer (AIME) as an answer, we report the exact match. 
% The implementation of this metric is taken directly from~\cite{lightman2023let}. 
The benchmark tool can also use LLM-as-a-judge to compare the model responses to the gold answers for cases where parsing is imperfect.
% Since parsing can be imperfect,  can use LLM-as-a-judge to %(GPT-5-mini~\cite{singh2025openai}), 
% compare the model responses to the gold answers.
%We ask GPT-5-mini~\cite{singh2025openai} to compare the LLM response with the gold answer.

% by prompting GPT 5-mini~\cite{singh2025openai} to evaluate the correctness of the generated answer by comparing it with the gold sample of the dataset.

% For HumanEval+ and MBPP+, we report the pass@1. Evaluation is performed using the \textit{EvalPlus} package, which runs each generated solution against a set of tests, comprising both the original base tests and the expanded set of test cases introduced by EvalPlus. A solution is considered correct only if it passes all tests.
For the coding benchmarks HumanEval+ and MBPP+, we report pass@1 accuracy. Solutions are evaluated using the \textit{EvalPlus} package; a generated response is classified as correct strictly when it successfully passes the original base tests alongside the expanded test cases introduced by EvalPlus.

For KernelBench, we use three evaluation measures: (\emph{i})~\emph{Syntax Check}, which verifies whether the generated code is syntactically valid, (\emph{ii})~\emph{Compilation Check}, which ensures that the code compiles successfully, and (\emph{iii})~\emph{Correctness Check}, which evaluates whether the generated kernel programming code produces outputs that are consistent with the native PyTorch implementation within a specified floating-point tolerance. We report the values for all evaluation measures as the proportion of successful cases.

\vspace{0.1cm}
\ecoparagraph{Efficiency metrics used:} (1) inference cost in TFLOPs \cite{kaplan2020scaling} and (2)~the number of generated tokens. The amount of TFLOPs is estimated according to \citet{hoffmann2022training}. 
We calculate the TFLOPs for prompt processing only once per input, reflecting KV cache reuse, while tracking generation costs across all individual samples. The computational overhead of running PRMs is also included in the total.
%We account for the fact that processing prompt tokens can leverage the KV cache, in contrast to generating new tokens in multiple different samples.
%We also include the TFLOPs associated with running PRMs. 
The computational overhead of information-theoretic uncertainty scorers (e.g., perplexity, sequence probability, and entropy) is considered negligible. A detailed description of metric computation is provided in Appendix~\ref{app:exp_details}.

% Note that TFLOPs estimates reflect \emph{theoretical} throughput and do not account for runtime optimizations such as vLLM prefix caching, which can substantially reduce wall-clock time by reusing KV-cache across answers that share a common prefix.
Note that our TFLOPs estimates reflect \emph{theoretical} compute costs and do not account for systems-level optimizations, such as vLLM prefix caching, which can substantially reduce wall-clock latency by reusing the KV cache across generations that share a common prefix.
We report theoretical TFLOPs to ensure reproducibility and hardware-independent comparisons.

%because they are computed from the token-level log-probabilities that the generator already produces during inference; no extra forward passes are needed. In contrast, PRM scoring requires a separate reward model to evaluate every candidate trajectory at every step, which empirically results in 5--10$\times$ higher total compute compared to uncertainty scorers on the same strategy. % TODO: add reference to Results section once compute trade-off analysis is written

\vspace{0.1cm}
\ecoparagraph{Details on scorer and strategy} configurations
(aggregation functions and scoring windows) are provided in Appendix~\ref{app:exp_details} and \ref{app:additional-results}.
% \ecoparagraph{Scorers.}
% We evaluate four scorer families: a PRM~\cite{yang2024qwen2}, uncertainty-based scorers (entropy, perplexity, sequence probability, pd\_gap), and ReProbe~\cite{ni2025efficient}. Details on scorer setup and hyperparameters for all strategies are in Appendix~\ref{app:additional-results} and~\ref{app:exp_details}.

% \ecoparagraph{Hyperparameters.}
% Hyperparameters for all strategies are listed in \Cref{app:exp_details}.

\subsection{Results}

% RQ: can we do better, if we have access to logits and hidden states.
% RQ: what is the best trade-off strategy

\ecoparagraph{Trade-off between performance and compute.}
The results for non-thinking Qwen2.5-Math-7B are presented in \Cref{fig:qwen25-ratio}, results for Qwen3-8B with native thinking on HumanEval-Plus in \Cref{fig:qwen3-humaneval-ratio}, and additional results on AIME can be found in \Cref{fig:qwen3-aime-ratio} in Appendix \ref{app:additional-results}. 
% On mathematical datasets, the best performance is achieved by PRMs scorers. However, on coding tasks (\Cref{fig:qwen3-humaneval-ratio}), the PRM does not consistently outperform lightweight uncertainty-based scorers. The configuration with uncertainty-based scorers and the MUR strategy substantially outperforms all baselines on HumanEval-Plus.
On mathematical benchmarks, the best results are obtained using PRM-based scorers. In contrast, on coding tasks (\Cref{fig:qwen3-humaneval-ratio}), PRMs do not consistently outperform lightweight uncertainty-based scorers. 

In particular, the combination of an uncertainty-based scorer with the MUR strategy substantially surpasses all baselines on HumanEval-Plus.
A plausible explanation is that the PRM in our experiments is trained predominantly on mathematical data and thus overfits to it. In contrast, uncertainty-based scorers are domain-agnostic and generalize more effectively to code generation, which constitutes an out-of-distribution setting for the PRM. 

Overall, uncertainty emerges as a robust and competitive scoring signal. Given their simplicity, near-zero computational overhead, and domain independence, uncertainty-based scorers offer a practical alternative in real-world deployments. However, our findings also underscore a clear need for coding-specific PRMs, a research area that remains largely underexplored. 
%Yet, our findings highlight the need for coding-specific PRMs, which remain largely underexplored as of now.

% This might be explained by the fact that PRM is overfit to mathematical problems, while uncertainty scores do not depend on this  domain so they perform better in the OOD scenario. Overall, uncertainty shows itself as a relatively effective scorer across all tasks.  Considering simplicity of uncertainty-based scorers, almost zero computational overhead, and domain independence they might be used in some practical scenarios. Our results also illustrate the need for coding-specific PRMs, which are currently scarce.

% First, we see that uncertainty can serve as an effective scorer. For both beam search and BoN, they provide substantial improvements over the baseline without scaling. For the HumanEval dataset with Qwen-3, it actually provides the best performance. Considering simplicity of uncertainty-based scorers and almost zero computational overhead, they might be used in some practical scenarios. On mathematical benchmarks, PRMs usually provide better results than uncertainty scores, while introducing substantial computational overhead. Notably, on coding tasks (\Cref{fig:qwen3-humaneval-ratio}), the PRM does not consistently outperform lightweight uncertainty-based scorers.

%since no code-specific PRM is currently available and we use a math-trained model (Qwen2.5-Math-PRM-7B) as a proxy, highlighting the need for domain-specific PRMs. 

Across TTC scaling strategies, beam search often underperforms compared to BoN and even the self-consistency baseline, despite requiring substantially more compute. Nevertheless, when it is paired with PRM-based scoring, it achieves the highest absolute performance on mathematical benchmarks. Dynamic TTC scaling (MUR) is a more efficient alternative to beam search and delivers the strongest results on HumanEval+. However, on mathematical datasets, it still lags behind BoN, even when combined with PRMs.

\vspace{0.6cm}
\ecoparagraph{Real world application: optimization of CUDA kernels and programming.}
\Cref{tab:gpt_oss} presents the results for coding tasks with GPT-OSS-120B. CUDA kernels generated with Offline BoN guided by a PRM have 5\% fewer syntax errors. The compilation rate is slightly lower as PRM tends to select more sophisticated code produced by the LLM, which can be more effective but also more prone to compilation failures. 
Importantly, the overall correctness is 4pp higher, i.e.,~\thinkbooster improves end-to-end CUDA kernel quality.

% Need better PRMs

% Present the experimental setup, datasets, evaluation metrics, and results demonstrating the effectiveness of ThinkBooster.

% \section{Conclusion}
% % Summarize the contributions of ThinkBooster and outline potential future work.

% To facilitate reproducible research and practical adoption.

\section{Conclusion}
%\todo{Make this fit in 6 pages; any word about Future Work?}

% Taken together, \thinkbooster provides an actionable pathway to make ``test-time compute scaling'' a controllable systems knob: instead of choosing a single expensive reasoning mode, practitioners can deploy a principled escalation strategy that uses inexpensive internal signals to spend compute only where it is likely to pay off.

We introduced \thinkbooster, a unified framework for TTC scaling of LLM reasoning that bridges research and deployment. By combining a modular library of TTC scaling strategies and scorers, a joint performance-compute benchmark, and an OpenAI-compatible endpoint that provides compute scaling as a service, \thinkbooster enables principled comparison and seamless integration of adaptive reasoning into real-world systems. Our results show that uncertainty can be used for both scoring and dynamic scaling of compute, achieving, in some cases, better results than PRMs and static methods.
%aware TTC scaling methods can achieve superior quality-cost trade-offs compared to static scaling.
We hope \thinkbooster will facilitate more systematic, compute-aware research and the practical adoption of test-time compute scaling.

\section*{Limitations}
Several components of our proposed \thinkbooster framework depend on deployment-specific capabilities. Some strategies and scorers require \emph{white-box} signals such as logits or hidden states, or API features such as prefill-style continuation, which are not available for fully hosted, black-box commercial models. \thinkbooster offers its full range of dynamic, uncertainty-driven strategies only against open-weight or self-hosted LLM; for closed APIs, it supports a black-box subset (best-of-$N$, majority voting, extended thinking with optional logits, and LLM-as-a-judge scoring). Reliable step-boundary extraction also remains challenging for large reasoning models with native, unstructured ``thinking'' traces, which can affect online scoring and escalation.

Our present empirical study has focused on a relatively narrow set of tasks -- primarily math, coding, and graduate-level scientific QA; therefore, the observed quality-cost trade-offs may not generalize to other settings such as long-context question answering, open-ended generation, or tool-augmented agents. For efficiency, we report \emph{theoretical} TFLOPs and number of generated tokens in order to ensure reproducibility and hardware-independent comparisons. We note that wall-clock latency is highly sensitive to batching, KV-cache reuse, hardware, and provider-side serving optimizations (e.g., vLLM prefix caching), and thus we leave a wall-clock study under a fixed serving configuration as a direction for future work and instead expose wall-clock measurement.
as a first-class field in the benchmark's per-request log. 
% Finally, proxy-based TTC scaling can introduce variable latency and provider-dependent cost behavior, and judge-based evaluation may add noise or bias to correctness estimates.

\section*{Acknowledgements}
We thank the anonymous reviewers for their feedback, which helped improve the paper.

\bibliography{custom}

\appendix
\clearpage

\section*{Appendix}

\section{Visual Debugger UI}
\label{app:demo-ui}

We provide a web-based Visual Debugger for inspecting test-time compute scaling strategies, sampled trajectories, and scoring signals. It exposes the search process step by step, including candidate generation, pruning, reranking, and final selection. \Cref{fig:debugger} 
% in the main paper 
shows the reasoning timeline, step inspector, and trajectory tree. The tool also provides a configuration interface for selecting models, strategies, and scorers; a screencast and live demo are available at the URLs in the footnotes on p.1.
% of the main paper.

\section{Details on the Experimental Setup}
\label{app:exp_details}

\begin{table*}
\centering
\footnotesize
\resizebox{0.8\textwidth}{!}{\begin{tabular}{ll|rrr|rrr|rrr}
\toprule
\textbf{Strategy} & \textbf{Scorer} & \multicolumn{3}{c|}{MATH-500} & \multicolumn{3}{c|}{OlympiadBench} & \multicolumn{3}{c}{Gaokao 2023 EN} \\
 &  & EM (\%) & $\Delta$ & TFLOPs & EM (\%) & $\Delta$ & TFLOPs & EM (\%) & $\Delta$ & TFLOPs \\
\midrule
Baseline & --- & 83.2 $\pm$ 0.0 & +0.0 & 5133.2 $\pm$ 0.0 & 39.3 $\pm$ 0.0 & +0.0 & 9683.7 $\pm$ 0.0 & 68.6 $\pm$ 0.0 & +0.0 & 4119.7 $\pm$ 0.0 \\
\midrule
Self-Consistency & --- & 86.4 $\pm$ 0.5 & +3.2 & 35680.2 $\pm$ 69.6 & 44.7 $\pm$ 0.6 & +5.4 & 69723.3 $\pm$ 464.0 & 72.4 $\pm$ 1.2 & +3.8 & 28817.1 $\pm$ 85.1 \\
\midrule
Offline BoN & prm (product, w=all) & 87.4 $\pm$ 0.3 & +4.2 & 80319.5 $\pm$ 740.9 & 45.9 $\pm$ 1.0 & +6.6 & 165575.6 $\pm$ 2015.1 & 74.4 $\pm$ 0.6 & +5.8 & 67125.3 $\pm$ 616.8 \\
Offline BoN & entropy (product, w=5) & 84.5 $\pm$ 0.4 & +1.3 & 37656.8 $\pm$ 375.8 & 40.7 $\pm$ 1.3 & +1.4 & 78834.1 $\pm$ 1023.1 & 71.6 $\pm$ 2.1 & +3.0 & 31761.6 $\pm$ 315.9 \\
Offline BoN & perplexity (mean, w=5) & 86.0 $\pm$ 0.5 & +2.8 & 37656.8 $\pm$ 375.8 & 42.6 $\pm$ 0.2 & +3.3 & 78834.1 $\pm$ 1023.1 & 72.7 $\pm$ 0.7 & +4.1 & 31761.6 $\pm$ 315.9 \\
Offline BoN & sequence\_prob (product, w=5) & 85.3 $\pm$ 0.1 & +2.1 & 37656.8 $\pm$ 375.8 & 42.5 $\pm$ 0.7 & +3.2 & 78834.1 $\pm$ 1023.1 & 70.3 $\pm$ 0.4 & +1.7 & 31761.6 $\pm$ 315.9 \\
Offline BoN & pd\_gap (mean, w=5) & 85.9 $\pm$ 0.6 & +2.7 & 37656.8 $\pm$ 375.8 & 42.1 $\pm$ 0.5 & +2.8 & 78834.1 $\pm$ 1023.1 & 72.4 $\pm$ 0.5 & +3.8 & 31761.6 $\pm$ 315.9 \\
\midrule
Beam Search (mean, w=5.0) & entropy & 84.2 $\pm$ 0.9 & +1.0 & 306729.1 $\pm$ 6750.5 & 40.3 $\pm$ 0.7 & +1.0 & 701502.5 $\pm$ 5301.5 & 68.4 $\pm$ 0.4 & -0.2 & 311218.7 $\pm$ 5769.7 \\
Beam Search (mean, w=5.0) & perplexity & 85.3 $\pm$ 0.7 & +2.1 & 303788.7 $\pm$ 6198.3 & 42.6 $\pm$ 1.2 & +3.3 & 701853.4 $\pm$ 6775.1 & 69.9 $\pm$ 0.3 & +1.3 & 297267.2 $\pm$ 2357.0 \\
Beam Search (mean, w=5.0) & sequence\_prob & 85.0 $\pm$ 0.7 & +1.8 & 375115.0 $\pm$ 13842.0 & 40.6 $\pm$ 1.7 & +1.3 & 925514.1 $\pm$ 18197.9 & 70.6 $\pm$ 1.0 & +2.0 & 373726.1 $\pm$ 2143.7 \\
Beam Search (mean, w=5.0) & uncertainty\_pd & 85.1 $\pm$ 0.9 & +1.9 & 310079.4 $\pm$ 2189.3 & 43.2 $\pm$ 1.0 & +3.9 & 704341.9 $\pm$ 24828.4 & 71.2 $\pm$ 0.4 & +2.6 & 299652.8 $\pm$ 5539.0 \\
\midrule
Beam Search (min, w=5.0) & entropy & 83.7 $\pm$ 1.1 & +0.5 & 322117.4 $\pm$ 10951.5 & 39.8 $\pm$ 1.0 & +0.5 & 736420.0 $\pm$ 6198.7 & 67.7 $\pm$ 0.5 & -0.9 & 320819.0 $\pm$ 6328.3 \\
Beam Search (min, w=5.0) & perplexity & 84.6 $\pm$ 1.6 & +1.4 & 319533.9 $\pm$ 10406.0 & 42.2 $\pm$ 0.1 & +2.9 & 730107.3 $\pm$ 21923.2 & 70.4 $\pm$ 1.1 & +1.8 & 327650.9 $\pm$ 7794.4 \\
Beam Search (min, w=5.0) & prm & 87.2 $\pm$ 0.6 & +4.0 & 1394300.9 $\pm$ 22014.1 & 47.6 $\pm$ 1.0 & +8.3 & 3957131.0 & 75.6 $\pm$ 0.7 & +7.0 & 1274337.7 \\
Beam Search (min, w=5.0) & sequence\_prob & 84.8 $\pm$ 1.2 & +1.6 & 379607.8 $\pm$ 6254.6 & 40.8 $\pm$ 0.9 & +1.5 & 968671.2 $\pm$ 5084.4 & 69.6 $\pm$ 0.8 & +1.0 & 381225.0 $\pm$ 3956.5 \\
Beam Search (min, w=5.0) & uncertainty\_pd & 84.7 $\pm$ 0.7 & +1.5 & 319126.0 $\pm$ 4238.3 & 40.9 $\pm$ 1.1 & +1.6 & 736593.5 $\pm$ 22199.8 & 71.9 $\pm$ 0.3 & +3.3 & 319073.4 $\pm$ 9783.5 \\
\midrule
Beam Search (mean, w=all) & reprobe & 85.0 $\pm$ 0.8 & +1.8 & 306865.7 & 41.8 $\pm$ 1.0 & +2.5 & 702565.9 & 72.2 $\pm$ 0.5 & +3.6 & 302712.9 \\
\midrule
MUR & entropy & 84.0 $\pm$ 0.5 & +0.8 & 49390.3 $\pm$ 1518.4 & 40.2 $\pm$ 0.9 & +0.9 & 91577.3 $\pm$ 1463.6 & 68.4 $\pm$ 0.9 & -0.2 & 39672.3 $\pm$ 1064.6 \\
MUR & perplexity & 84.5 $\pm$ 0.1 & +1.3 & 52408.6 $\pm$ 4448.2 & 40.3 $\pm$ 1.5 & +1.0 & 105190.2 $\pm$ 2220.2 & 68.6 $\pm$ 0.9 & +0.0 & 47704.4 $\pm$ 576.4 \\
MUR & prm & 84.0 $\pm$ 0.5 & +0.8 & 169527.8 $\pm$ 2908.5 & 42.7 $\pm$ 0.7 & +3.4 & 431315.4 $\pm$ 27690.9 & 70.6 $\pm$ 0.8 & +2.0 & 138831.0 $\pm$ 4931.2 \\
MUR & sequence\_prob & 83.6 $\pm$ 0.5 & +0.4 & 54883.8 $\pm$ 2568.1 & 40.4 $\pm$ 1.1 & +1.1 & 120399.4 $\pm$ 6429.8 & 68.8 $\pm$ 1.6 & +0.2 & 44550.5 $\pm$ 1744.8 \\
MUR & uncertainty\_pd & 83.8 $\pm$ 0.2 & +0.6 & 48814.0 $\pm$ 1998.4 & 40.2 $\pm$ 1.3 & +0.9 & 97544.3 $\pm$ 2685.9 & 68.4 $\pm$ 0.8 & -0.2 & 42118.9 $\pm$ 1106.7 \\
\bottomrule
\end{tabular}}
\caption{Full results for Qwen2.5-Math-7B-Instruct across mathematical benchmarks. For each strategy--scorer pair, we report exact match (EM) accuracy together with standard deviation, improvement over the baseline ($\Delta$), as well as compute cost in terms of TFLOPs.}
\label{tab:qwen25-full-results}
\end{table*}

\begin{table*}[!ht]
\centering
\footnotesize
\resizebox{\textwidth}{!}{\begin{tabular}{ll|rrr|rrr|rrr}
\toprule
\textbf{Strategy} & \textbf{Scorer} & \multicolumn{3}{c|}{AIME 2024} & \multicolumn{3}{c|}{AIME 2025} & \multicolumn{3}{c}{HumanEval-Plus} \\
 &  & EM (\%) & $\Delta$ & TFLOPs & EM (\%) & $\Delta$ & TFLOPs & Score (\%) & $\Delta$ & TFLOPs \\
\midrule
Baseline & --- & 75.6 $\pm$ 2.7 & 0.0 & 6626.9 $\pm$ 153.1 & 64.4 $\pm$ 6.2 & 0.0 & 8083.6 $\pm$ 305.2 & 79.3 $\pm$ 0.6 & 0.0 & 10094.9 $\pm$ 468.1 \\
\midrule
Extended Thinking & --- & 78.9 $\pm$ 5.1 & 3.3 & 26477.7 $\pm$ 1594.4 & 66.7 $\pm$ 3.3 & 2.3 & 27573.9 $\pm$ 1140.5 & 80.3 $\pm$ 2.3 & 1.0 & 36625.5 $\pm$ 504.3 \\
\midrule
Self-Consistency & --- & 82.2 $\pm$ 1.9 & 6.6 & 52797.2 $\pm$ 816.3 & 73.3 $\pm$ 5.8 & 8.9 & 53684.9 $\pm$ 1499.8 & --- & --- & --- \\
\midrule
Offline BoN & prm (min, w=15) & 81.1 $\pm$ 3.8 & 5.5 & 65399.1 $\pm$ 421.2 & 74.4 $\pm$ 3.9 & 10.0 & 76286.1 $\pm$ 350.1 & 80.9 $\pm$ 0.9 & 1.6 & 120938.7 $\pm$ 1038.0 \\
Offline BoN & entropy (max, w=all) & 74.4 $\pm$ 5.1 & -1.2 & 52455.6 $\pm$ 392.2 & 70.0 $\pm$ 3.3 & 5.6 & 63349.7 $\pm$ 286.3 & 80.7 $\pm$ 3.4 & 1.4 & 73546.4 $\pm$ 950.6 \\
Offline BoN & perplexity (max, w=all) & 72.2 $\pm$ 6.9 & -3.4 & 52455.6 $\pm$ 392.2 & 68.9 $\pm$ 5.1 & 4.5 & 63349.7 $\pm$ 286.3 & 80.9 $\pm$ 1.5 & 1.6 & 73546.4 $\pm$ 950.6 \\
Offline BoN & sequence\_prob (mean, w=1) & 75.6 $\pm$ 1.9 & 0.0 & 52455.6 $\pm$ 392.2 & 74.4 $\pm$ 3.9 & 10.0 & 63349.7 $\pm$ 286.3 & 80.3 $\pm$ 0.7 & 1.0 & 73546.4 $\pm$ 950.6 \\
Offline BoN & pd\_gap (max, w=all) & 75.6 $\pm$ 1.9 & 0.0 & 52455.6 $\pm$ 392.2 & 67.8 $\pm$ 6.9 & 3.4 & 63349.7 $\pm$ 286.3 & 81.7 $\pm$ 1.6 & 2.4 & 73546.4 $\pm$ 950.6 \\
\midrule
Beam Search (min, w=5.0) & entropy & 75.6 $\pm$ 1.9 & 0.0 & 658505.8 $\pm$ 89881.0 & 62.2 $\pm$ 3.8 & -2.2 & 896577.1 $\pm$ 52575.6 & 84.1 $\pm$ 1.6 & 4.8 & 973513.9 $\pm$ 47317.0 \\
Beam Search (min, w=5.0) & perplexity & 72.2 $\pm$ 3.8 & -3.4 & 745916.4 $\pm$ 85315.2 & 64.4 $\pm$ 7.7 & 0.0 & 868007.9 $\pm$ 14121.5 & 81.9 $\pm$ 0.7 & 2.6 & 1061892.2 $\pm$ 101885.9 \\
Beam Search (min, w=5.0) & prm & 71.1 $\pm$ 1.9 & -4.5 & 3294504.2 $\pm$ 204100.1 & 66.7 $\pm$ 9.4 & 2.3 & 3625247.2 $\pm$ 285892.8 & 84.8 $\pm$ 1.8 & 5.5 & 3157463.9 $\pm$ 265871.8 \\
Beam Search (min, w=5.0) & sequence\_prob & 68.9 $\pm$ 5.1 & -6.7 & 1809340.9 $\pm$ 122115.7 & 61.1 $\pm$ 5.1 & -3.3 & 2528486.9 $\pm$ 148198.6 & 81.7 $\pm$ 2.8 & 2.4 & 1287438.5 $\pm$ 211202.5 \\
Beam Search (min, w=5.0) & pd\_gap & 66.7 $\pm$ 3.3 & -8.9 & 777367.6 $\pm$ 128986.7 & 65.6 $\pm$ 8.4 & 1.2 & 976551.8 $\pm$ 80244.4 & 82.9 $\pm$ 0.0 & 3.6 & 932738.7 $\pm$ 113300.2 \\
\midrule
MUR & entropy & 74.4 $\pm$ 3.8 & -1.2 & 313539.7 $\pm$ 45205.0 & 65.6 $\pm$ 5.1 & 1.2 & 457755.2 $\pm$ 20973.2 & 88.8 $\pm$ 0.9 & 9.5 & 723296.3 $\pm$ 17034.3 \\
MUR & perplexity & 76.7 $\pm$ 0.0 & 1.1 & 312986.7 $\pm$ 13905.3 & 68.9 $\pm$ 5.1 & 4.5 & 320316.1 $\pm$ 9509.8 & 87.0 $\pm$ 1.9 & 7.7 & 764281.0 $\pm$ 65294.1 \\
MUR & prm & 77.8 $\pm$ 3.8 & 2.2 & 1446222.0 $\pm$ 129645.7 & 67.8 $\pm$ 5.1 & 3.4 & 968297.5 $\pm$ 66626.3 & 82.1 $\pm$ 3.6 & 2.8 & 945664.2 $\pm$ 575351.3 \\
MUR & sequence\_prob & 76.7 $\pm$ 3.3 & 1.1 & 340592.8 $\pm$ 42368.6 & 70.0 $\pm$ 5.8 & 5.6 & 459197.6 $\pm$ 148899.7 & 87.6 $\pm$ 0.9 & 8.3 & 689156.8 $\pm$ 14247.9 \\
MUR & pd\_gap & 78.9 $\pm$ 5.1 & 3.3 & 312766.3 $\pm$ 18273.8 & 65.6 $\pm$ 3.8 & 1.2 & 531023.4 $\pm$ 41281.8 & 86.0 $\pm$ 1.6 & 6.7 & 758145.9 $\pm$ 51357.1 \\
\bottomrule
\end{tabular}}
\caption{Full results for Qwen3-8B across mathematical and coding benchmarks. For each strategy--scorer pair, we report accuracy together with standard deviation, improvement over the baseline ($\Delta$), as well as compute cost in terms of TFLOPs.}
\label{tab:qwen3-full-results}
\end{table*}

\paragraph{Model--dataset mapping.}
Qwen2.5-Math-7B is evaluated on three mathematical datasets (MATH-500, OlympiadBench, Gaokao~2023~EN). Qwen3-8B is evaluated on mathematical (AIME~2024, AIME~2025) and coding (HumanEval+) tasks. GPT-OSS-120B is evaluated on scientific QA (GPQA-Diamond) and coding tasks (HumanEval+, MBPP+, KernelBench). All experiments use vLLM~\cite{kwon2023efficient} as the inference backend. Results are averaged over 3 seeds with the standard deviation reported where applicable.

% \paragraph{Hyperparameters.}
% Generation parameters differ by model.
% For \textbf{Qwen2.5-Math-7B}: temperature~$=0.7$, top-$p$~$=0.8$, top-$k$~$=20$, max tokens~$=4096$.
% For \textbf{Qwen3-8B}: temperature~$=0.6$, top-$p$~$=0.95$, top-$k$~$=20$, max tokens~$=32768$ (native thinking mode enabled).
% For \textbf{GPT-OSS-120B}: temperature~$=0.6$, top-$p$~$=0.95$, top-$k$~$=20$, max tokens~$=65536$.
% Strategy-specific settings:
% beam search uses beam size~$=5$, candidates per beam~$=8$, max steps~$=30$ for Qwen2.5, and beam size~$=3$, candidates per beam~$=5$, max steps~$=250$ for Qwen3.
% For offline BoN and self-consistency: $N{=}8$ samples.
% For MUR: candidates per step~$=8$, momentum rate~$=0.9$, max steps~$=50$ (Qwen2.5) / $250$ (Qwen3).
% For extended thinking (Qwen3 only): max continuations~$=3$.

\paragraph{Hyperparameters} vary by model.
\textbf{Qwen2.5-Math-7B}: temperature 0.7, top-$p$ 0.8, top-$k$ 20, max tokens 4096.
\textbf{Qwen3-8B}: temperature 0.6, top-$p$ 0.95, top-$k$ 20, max tokens 32768, with native thinking mode enabled.
\textbf{GPT-OSS-120B}: temperature 0.6, top-$p$ 0.95, top-$k$ 20, max tokens 65536.

We use beam size 5 with 8 candidates, and 30 max steps for Qwen2.5; and beam size 3 with 5 candidates, and 250 max steps for Qwen3. Offline BoN and self-consistency use 8 samples. MUR uses 8 candidates per step, momentum 0.9, and up to 50 steps for Qwen2.5 or 250 for Qwen3. Extended thinking for Qwen3 allows up to 3 continuations.

\paragraph{Datasets.}
\Cref{tab:datasets} gives statistics for each dataset in our experiments.
MATH-500 is a representative subset of the MATH benchmark~\cite{math500} used for symbolic or numeric answer checking. OlympiadBench~\cite{he2024olympiadbench} is an olympiad-level math and physics benchmark. % Minerva Math is a quantitative/STEM reasoning benchmark from the Minerva work~\cite{lewkowycz2022solving}.
GaoKao23EN is an English math QA compilation centered on 2023 exam-style items~\cite{zhang2023evaluating}.
AIME-2024 and AIME-2025~\cite{aime_problems_solutions} are problems from the American Invitational Mathematics Examination.

For \emph{scientific QA}, we use GPQA-Diamond~\cite{rein2023gpqa}, a subset of GPQA with graduate-level MCQs in biology, chemistry, and physics that require expert-level reasoning.

For \emph{coding}, we use HumanEval+ and MBPP+~\cite{liu2023your}, which are extensions of HumanEval~\cite{chen2021evaluating} and MBPP~\cite{austin2021program}. MBPP+ consists of entry-level Python tasks, while HumanEval+ contains moderately challenging function-level Python problems.
KernelBench~\cite{ouyang2025kernelbench} poses the real-world problem of writing GPU kernels: producing CUDA kernel code that outperforms PyTorch's native implementations. We focus on foundational Level-1 operators.

\begin{table}
\footnotesize
\centering
\resizebox{0.43\textwidth}{!}{%
\begin{tabular}{l|l|c}
\toprule
\textbf{Task} & \textbf{Dataset} & \textbf{\# test samples} \\
\midrule
\multirow{5}{*}{Math} 
& MATH-500 & 500 \\
& OlympiadBench & 675 \\
%& Minerva Math & 272 \\
& GaoKao23EN & 385 \\
& AIME-2024 & 30 \\
& AIME-2025 & 30 \\
\midrule
Scientific QA 
& GPQA-Diamond & 198 \\
\midrule
\multirow{3}{*}{Coding} 
& HumanEval+ & 164 \\
& MBPP+ & 378 \\
& KernelBench & 100 \\
\bottomrule
\end{tabular}%
}
\caption{\label{tab:datasets} Statistics of the datasets used in the experiments.}
\end{table}

\paragraph{TFLOPs metric.} For each forward pass, we compute $\text{FLOPs} = 2 \times N \times T$, where $N$ is the number of model parameters and $T$ is the total number of tokens (input context plus generated output). This approximation counts one multiply-add per weight per token, and is standard in scaling-law analysis~\cite{hoffmann2022training}.
% across configurations.
We track token counts separately for the reasoning generator and, when applicable, for PRMs. The generator tracker records the context tokens once per prompt, even when the LLM generates multiple candidates from a shared prefix, and output tokens summed over all candidates. The PRM tracker records the number of input tokens for each scoring call; because PRM is a reward model that only performs a forward pass without generation, its cost is $2 \times N_{\text{PRM}} \times T_{\text{input}}$. Finally, $\text{TFLOPs}_{\text{total}} = \text{TFLOPs}_{\text{gen}} + \text{TFLOPs}_{\text{PRM}}$.
The overhead for computing uncertainty-based scorers is considered negligible.% (entropy, perplexity, sequence probability) 

\paragraph{Reasoning step extraction.}
% All models receive a simple system prompt (e.g., ``Please reason step by step, and put your final answer within \textbackslash boxed\{\}'') without any explicit formatting instructions. Step boundaries are detected \emph{post-hoc} by our \texttt{ThinkingMarkerDetector}, which uses linguistic markers across eight categories: sequence indicators (``first'', ``next'', ``then''), conclusion signals (``therefore'', ``thus'', ``hence''), thinking cues (``let me'', ``wait'', ``actually''), verification phrases (``to check'', ``substituting''), reasoning connectives (``note that'', ``consider''), sentence-start transitions (``but'', ``however'', ``since''), self-correction signals (``mistake'', ``wrong''), and structural patterns (paragraph breaks, numbered lists). A split is triggered only at sentence boundaries (after~\texttt{.!?\textbackslash n}) to avoid mid-sentence fragmentation. Steps are then normalized by merging fragments below a minimum token count and splitting oversized segments.

All models are prompted with a simple instruction, for example, ``\emph{Please reason step by step and put your final answer within \textbackslash boxed\{\}}'', without additional formatting constraints. Step boundaries are detected post hoc using a \texttt{ThinkingMarkerDetector} that identifies linguistic cues across categories such as sequence indicators, conclusion signals, verification phrases, self-corrections, and structural patterns. 

Splits occur only at sentence boundaries to avoid mid-sentence breaks. The resulting segments are then normalized by merging short fragments and splitting overly long ones.
For thinking-mode models (Qwen3-8B), the same detector operates on the content inside \verb|<think></think>| tags.

\section{Additional Experimental Results}
\label{app:additional-results}

\paragraph{Scorer details.}
As the PRM scorer, we use Qwen2.5-Math-PRM-7B~\cite{yang2024qwen2}, a process reward model trained on mathematical reasoning data and deployed on a separate GPU. Since no coding PRM is released, we use the same math-trained PRM as a proxy for coding tasks; the scorer abstraction supports per-domain PRM swap via a config flag.
For uncertainty-based scorers, we use entropy, perplexity, sequence probability, and probability differential, all computed on token-level log-probabilities produced during generation with negligible overhead.
We also evaluate ReProbe~\cite{ni2025efficient}, a lightweight linear probe ($<$10M~parameters) trained on internal model representations.

% \paragraph{Scorer configurations.}
% For each strategy--scorer combination in \Cref{fig:accuracy-vs-compute,fig:qwen3-aime-ratio}, we select the best step-level aggregation function and scoring window via grid search.
% For \textbf{offline BoN}, the PRM scorer uses product aggregation over all reasoning steps ($w{=}\text{all}$), while the lightweight uncertainty-based scorers (entropy, perplexity, sequence probability, probability-differential gap) use a window of $w{=}5$ steps with their respectively best aggregation (product or mean).
% For \textbf{beam search}, we report both mean and min step-level aggregation with $w{=}5$; additionally, we include beam search guided by the ReProbe scorer~\cite{ni2025efficient} using mean aggregation over all steps ($w{=}\text{all}$). The ReProbe scorer is a lightweight linear probe head ($<$10M parameters) trained on internal model representations, adding negligible compute overhead (${\sim}$0.15\% of the base model).
% \textbf{MUR} uses each scorer with its default configuration.
% Compute is measured in TFLOPs and includes both generation and scoring costs.

\begin{figure}
    \centering
    \includegraphics[width=\linewidth]{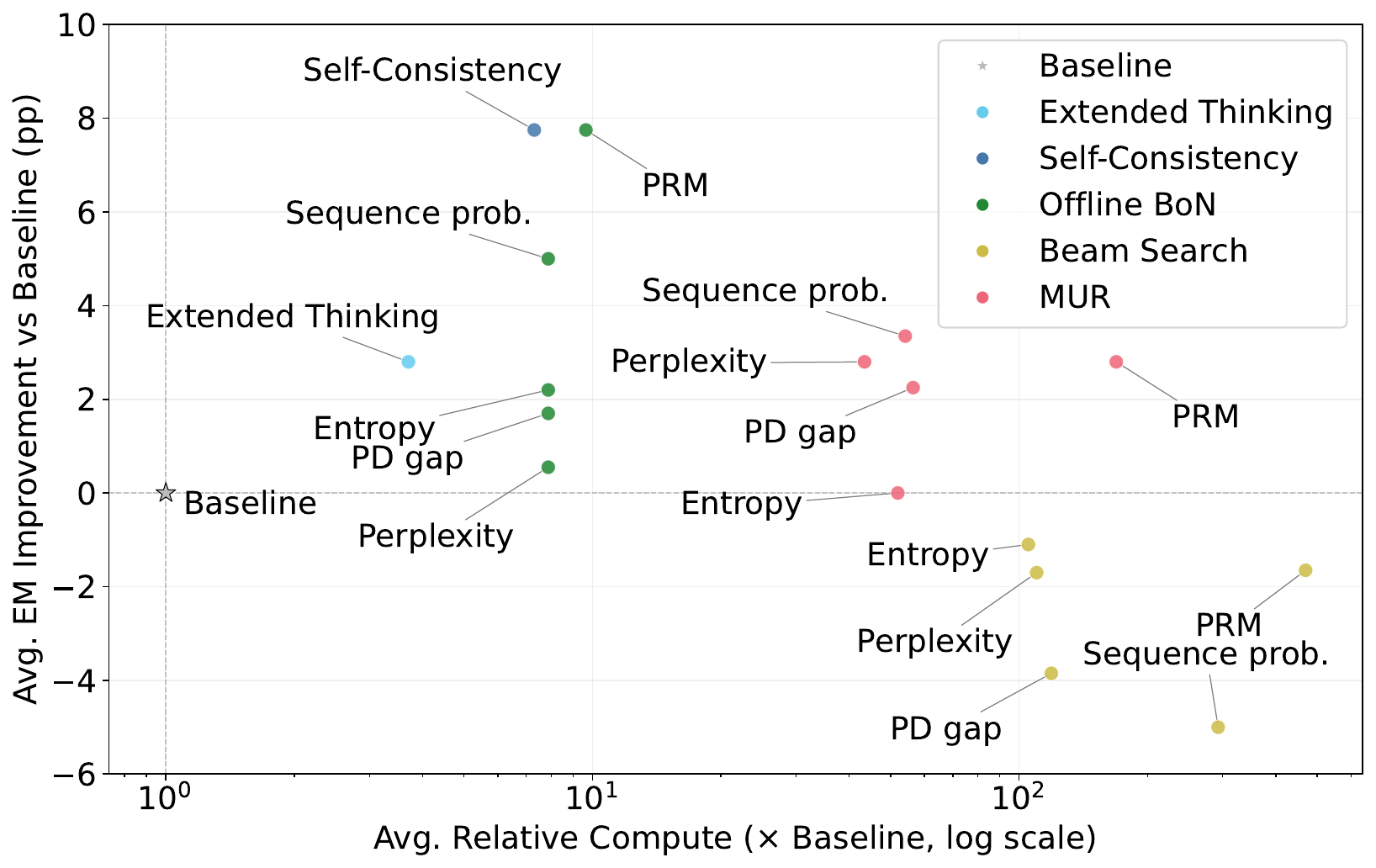}
    \caption{Accuracy improvement vs.\ compute ratio for Qwen3-8B on AIME (aggregate of AIME 2024 and AIME 2025). Each point represents a strategy--scorer combination; the $x$-axis shows the compute ratio relative to the baseline (log scale).}
    \label{fig:qwen3-aime-ratio}
\end{figure}

\paragraph{Scorer configurations.}
% For the results reported in \Cref{fig:accuracy-vs-compute} in the main text and for \Cref{fig:qwen3-aime-ratio} here, we select the best step-level aggregation and scoring window via grid search.

For the results reported in \Cref{fig:accuracy-vs-compute} in the main text and in \Cref{fig:qwen3-aime-ratio}, we select the best combination of step-level aggregation function and scoring window through a grid search over the corresponding hyperparameter space. This is done independently for each scorer and strategy configuration for a fair comparison across methods.

For \textbf{offline BoN}, PRM uses product aggregation over all steps ($w{=}\text{all}$), while other scorers use a $w{=}5$ window with their best aggregation.

For \textbf{beam search}, we report the mean and the minimum aggregation with $w{=}5$, and we further include ReProbe~\cite{ni2025efficient} with mean aggregation over all steps. ReProbe is a lightweight linear probe head with negligible overhead.

\textbf{MUR} uses default settings. Compute is measured in TFLOPs for generation plus scoring.

\paragraph{Detailed per-dataset results.}
\Cref{tab:qwen25-full-results} reports the full per-dataset results for Qwen2.5-Math-7B-Instruct across mathematical benchmarks (MATH-500, OlympiadBench, GaoKao~2023~EN), shown in aggregate in \Cref{fig:qwen25-ratio}.
\Cref{tab:qwen3-full-results} reports the full per-dataset results for Qwen3-8B across the mathematical (AIME 2024, AIME 2025) and coding (HumanEval+) benchmarks, shown in aggregate in \Cref{fig:qwen3-humaneval-ratio} and \Cref{fig:qwen3-aime-ratio}. For each strategy--scorer pair, we report accuracy with standard deviation, improvement over the baseline ($\Delta$), as well as compute cost in TFLOPs.

\twocolumn

\end{document}